\documentclass{article}

\usepackage{microtype}
\usepackage{graphicx}
\usepackage{subcaption}
\usepackage{booktabs} % for professional tables

\usepackage{hyperref}

\usepackage[accepted]{icml2026}

\usepackage{amsmath}
\usepackage{amssymb}
\usepackage{mathtools}
\usepackage{amsthm}
\usepackage{xcolor}
\usepackage[table]{xcolor}
\usepackage{algorithm}
\usepackage{algorithmic}

\usepackage[capitalize,noabbrev]{cleveref}

\theoremstyle{plain}

\theoremstyle{definition}

\theoremstyle{remark}

\usepackage[textsize=tiny]{todonotes}

\icmltitlerunning{Token Sparse Attention: Efficient Long-Context Inference with Interleaved Token Selection}

\begin{document}

\twocolumn[
  \icmltitle{Token Sparse Attention: Efficient Long-Context Inference\\with Interleaved Token Selection}



  \icmlsetsymbol{equal}{*}

  \begin{icmlauthorlist}
  \icmlauthor{Dongwon Jo}{snu}
  \icmlauthor{Beomseok Kang}{snu}
  \icmlauthor{Jiwon Song}{snu}
  \icmlauthor{Jae-Joon Kim}{snu}
  \url{https://github.com/dongwonjo/Token-Sparse-Attention}
  \end{icmlauthorlist}

  \icmlaffiliation{snu}{Department of Electrical and Computer Engineering, Seoul National University, Seoul, South Korea}
  \icmlcorrespondingauthor{Beomseok Kang, Jae-Joon Kim}{beomseok, kimjaejoon@snu.ac.kr}
  \icmlkeywords{Machine Learning, ICML}

  \vskip 0.3in
]



\printAffiliationsAndNotice{}  % no special notice (required even if empty)

\begin{abstract}
The quadratic complexity of attention remains the central bottleneck in long-context inference for large language models.
Prior acceleration methods either sparsify the attention map with structured patterns or permanently evict tokens at specific layers, which can retain irrelevant tokens or rely on irreversible early decisions despite the layer-/head-wise dynamics of token importance.
In this paper, we propose Token Sparse Attention, a lightweight and dynamic token-level sparsification mechanism that compresses per-head $Q, K, V$ to a reduced token set during attention and then decompresses the output back to the original sequence, enabling token information to be reconsidered in subsequent layers.
Furthermore, Token Sparse Attention exposes a new design point at the intersection of token selection and sparse attention. Our approach is fully compatible with dense attention implementations, including Flash Attention, and can be seamlessly composed with existing sparse attention kernels.
Experimental results show that Token Sparse Attention consistently improves accuracy–latency trade-off, achieving up to $\times$3.23 attention speedup at 128K context with less than 1\% accuracy degradation. These results demonstrate that dynamic and interleaved token-level sparsification is a complementary and effective strategy for scalable long-context inference.
\end{abstract}

\begin{figure}[h]
    \centering
    \includegraphics[width=0.80\linewidth]{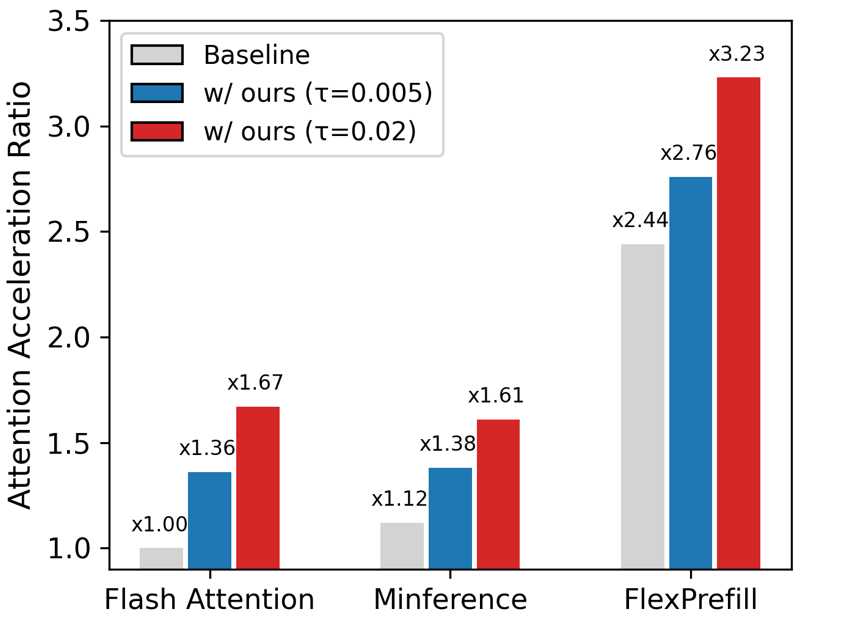}
     \vspace{-2pt}
    \caption{\textbf{Speedups with Token Sparse Attention.}
    Attention acceleration ratios obtained by applying the proposed Token Sparse Attention (ours) to existing attention acceleration methods. $\tau$ denotes the sparsity level (\textit{i.e.}, higher the sparser).
    }
    
    \vspace{-18pt}
    \label{fig:introduction}
\end{figure}

\begin{figure*}[t]
    \centering
    \includegraphics[width=1.0\linewidth]{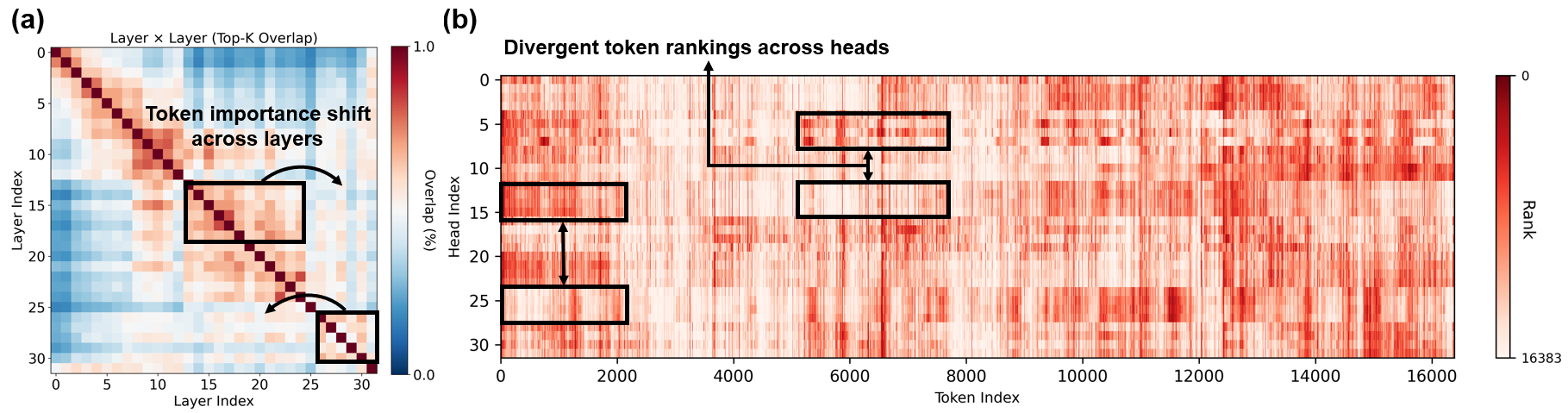}
     \vspace{-12pt}
    \caption{\textbf{Dynamics of Token Importance.}
    \textbf{(a)} Layer-wise overlap of top-k important tokens, showing that token importance shifts significantly across layers. \textbf{(b)} Head-wise token importance rankings within the same layer, illustrating that different attention heads prioritize different subsets of tokens.
    }
    \vspace{-12pt}
    \label{fig:motivation}
\end{figure*}

\section{Introduction}
The capability to process long contexts has become a defining feature of modern Large Language Models (LLMs)~\cite{gpt4, gemini, claude3}, enabling applications ranging from long document summarization to multi-turn reasoning and code generation~\cite{intro1,intro2,intro3}. However, as context lengths ($L$) increase, the complexity of attention mechanism grows quadratically ($O(L^{2})$) during the prefill stage, creating a fundamental bottleneck in inference. While hardware-aware optimizations such as FlashAttention~\cite{flashattention2} significantly reduce memory I/O overhead, the inherent quadratic complexity remains the same, necessitating algorithmic innovations to scale practical context lengths beyond 100K.

Recent state-of-the-art (SOTA) methods for accelerating prefill predominantly rely on \textit{sparse attention}~\cite{flexprefill, minference}, where computations for low-importance regions in the attention map ($QK^{\top}$) are bypassed. While these methods approximate attention weights effectively, they typically impose block-level sparsity to stay compatible with the execution patterns of attention kernels. As a result, less informative tokens may still be preserved when grouped with salient tokens in the same block, potentially leading to sub-optimal sparsity. A few studies have explored token-level sparsity during prefill~\cite{fastkv, gemfilter}. These methods identify a subset of important tokens in the early layers and permanently discard the rest in deeper layers. However, once tokens are evicted, they cannot be reconsidered even if they become significant in subsequent layers. This naturally enforces the same set of tokens to persist across layers and fails to capture more selective patterns such as layer- or/and head-specific shifts in token importance. Consequently, eviction-based approaches often exhibit weaker accuracy-speedup trade-offs, falling short of fully leveraging the benefits of operating at a finer (token-level) granularity.

In this paper, we introduce \textbf{Token Sparse Attention}, a novel token-level attention mechanism for accelerating prefill in long-context inference. At each head, the method selects a small subset of informative tokens ($L'{<}L$) and performs attention efficiently within the resulting compressed space ($\mathbb{R}^{L' \times L'}$), allowing the selected subset to differ across heads. However, naïvely applying this selection prevents the remaining tokens ($L{-}L'$) from being re-visited in subsequent layers, inevitably evicting them. To address this issue, we interleave the attention output ($\mathbb{R}^{L' \times d}$) back into the original sequence dimension ($\mathbb{R}^{L \times d}$). This “Compress and then Decompress” design enables each head to repeatedly select its own significant tokens from the full sequence ($L$), reaping the benefits of head-specific token selection. In addition, because the full sequence dimension is restored ($L'{\rightarrow}L$) at every layer, our design naturally supports layer-wise adaptive sparsity budgets. Consequently, our method dynamically determines the sparsity level on-the-fly during inference, providing a flexibility that eviction-based approaches inherently lack.

While improving performance on its own, Token Sparse Attention's core strength lies in being complementary to existing sparse attention methods rather than replacing them. This arises from its strong compatibility: irrelevant tokens can be pruned prior to any sparse computations, \textit{e.g.}, block-sparse or $\Lambda$-shape. Because our method essentially performs dense attention within the compressed space ($\mathbb{R}^{L' \times L'}$), it integrates seamlessly with implementations like FlashAttention and sparse attention kernels. As a result, we demonstrate heterogeneous granularity (see Figure~\ref{fig:introduction}) as an effective new strategy for enhancing sparse attention, \textit{e.g.}, combining FlexPrefill~\cite{flexprefill} (block-sparse) with ours (token-sparse) reaches 87.3\% accuracy with $\times$2.8 speedup than FlashAttention, whereas standard FlexPrefill achieves the same accuracy (87.3\%) but $\times$2.4 speedup (see Table~\ref{tab:ruler}).

\section{Method}

\subsection{Motivation}
\label{sec:motivation}

\begin{figure*}[t]
    \centering
    \includegraphics[width=1.0\linewidth]{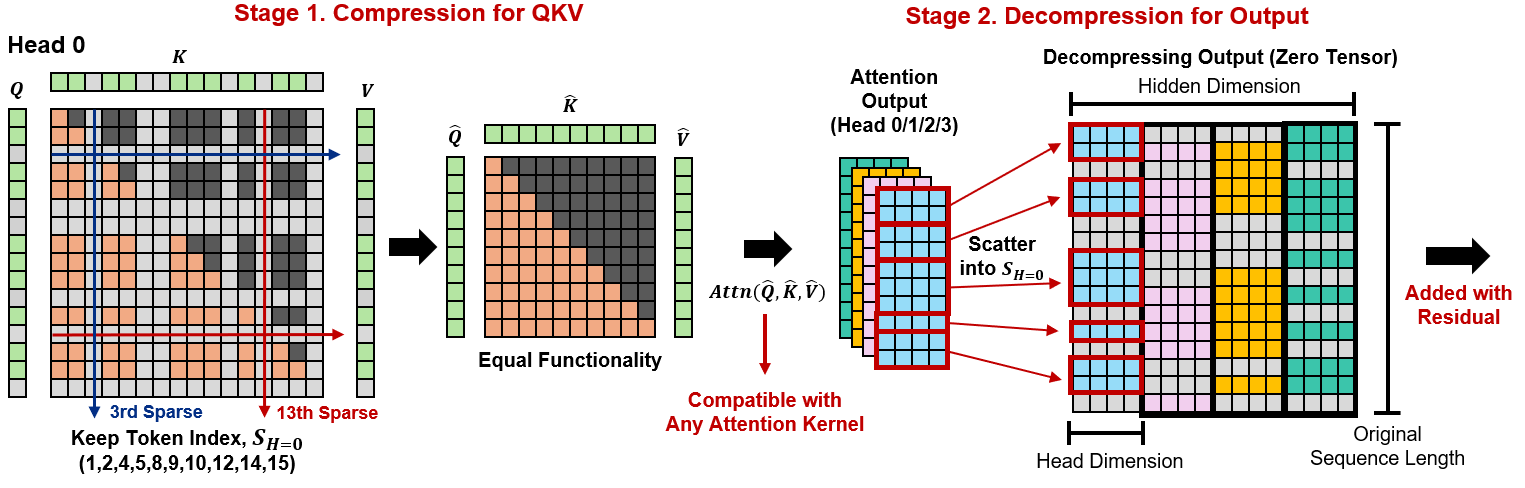}
     \vspace{-12pt}
    \caption{\textbf{Overview of the proposed Token Sparse Attention.} Stage 1 compresses $Q$, $K$, and $V$ by selecting a per-head token subset $S_{H=0}$, yielding compact $\hat{Q}$, $\hat{K}$, and $\hat{V}$ that remain compatible with standard attention kernels. Stage 2 performs attention on the compressed tensors and scatters the resulting outputs back into the full sequence layout before adding the residual connection.
    }
    \vspace{-10pt}
    \label{fig:token_sparse_attention}
\end{figure*}

Prior token-sparse techniques~\cite{fastkv, gemfilter} generally reduce the computational cost of attention by selecting a subset of tokens in early layers and restricting computation to only those tokens in subsequent layers. However, this strategy implicitly assumes that token importance estimated in early layers will remain stable throughout the model during prefill, which rarely holds in practice. To illustrate this, we analyze token-importance dynamics during prefill for a LLaMA-3.1-8B-Instruct model~\cite{llama3}, as shown in Figure~\ref{fig:motivation}.

\textbf{Layer-wise Token Importance.} Figure~\ref{fig:motivation}\textbf{(a)} reports the layer-wise overlap of important tokens, where we select the top 1$\%$ tokens at each layer and measure their pairwise overlap across layers. Although adjacent layers share a moderate fraction of important tokens, the overlap decreases rapidly as the layer distance increases. This indicates that token importance shifts substantially across layers as attention focus evolves during prefill. As a result, permanently removing tokens based on early-layer selection can prematurely eliminate candidates for subsequent layers that might later become relevant.

\textbf{Head-wise Token Importance.} Figure~\ref{fig:motivation}\textbf{(b)} further shows token-importance variation across attention heads at layer 18. Different heads exhibit distinct token-ranking patterns, suggesting that token relevance is not only layer-dependent but also head-specific. This heterogeneity is a fundamental property of multi-head attention, where heads specialize in capturing different contextual relationships. Thus, enforcing a coarse, layer-level eviction policy forces all heads to share a unified token set, inevitably discarding tokens that individual heads may find essential.

These observations indicate that token selection should retain flexibility across both layers and heads, rather than permanently committing to early-layer decisions. Motivated by this, Token Sparse Attention is designed to accommodate these dynamics through interleaved token selection, which restores the full sequence dimension after each compressed attention step. This reversible design reduces unnecessary computation while preserving future selection space.

\subsection{Token Sparse Attention}

Motivated by the observations in Section~\ref{sec:motivation}, we aim to selectively skip irrelevant tokens during attention computation without permanently removing them from the sequence.
As illustrated in Figure~\ref{fig:token_sparse_attention}, this process consists of two primary stages which are compression for query-key-value tensors and decompression for the attention output. These stages ensure that the computational benefits of sparsity are realized while maintaining the structural integrity of the original sequence.

\textbf{Stage 1: Compression for QKV.} In the compression stage, each attention head $h$ selects a subset of token indices $S_{H=h}$, yielding a reduced sequence length $L' \ll L$.
Using these indices, we gather the corresponding rows from the original $Q$, $K$, and $V$ tensors to construct compressed tensors $\hat{Q}$, $\hat{K}$, and $\hat{V}$.
Importantly, this selection is performed independently per head, allowing different heads to attend to different subsets of tokens. This design directly addresses the head-wise heterogeneity observed in Figure~\ref{fig:motivation}\textbf{(b)}.
A major advantage of this compression strategy is that the resulting tensors $\hat{Q}, \hat{K}, \hat{V}$ remain dense and contiguous in memory. This allows us to utilize highly optimized hardware-aware kernels such as FlashAttention or other specialized sparse attention implementations without any modification.
The attention operation is performed on the compressed tensors to produce a reduced output tensor $\hat{O}$, which contains the context-aware representations for the selected tokens, thereby reducing the quadratic attention cost from $O(L^{2}d)$ to $O(L'^{2}d)$.

\textbf{Stage 2: Decompression for Attention Output.} In the decompression stage following attention computation, the compressed output $\hat{O}$ is scattered back into a zero-initialized tensor of shape $\mathbb{R}^{L \times d}$ using the head-specific index set $S_{H=h}$.
This operation ensures that the output dimensions match the original input, preventing any dimension mismatch issues in subsequent layers. The unselected positions in the output tensor remain zero, which is functionally equivalent to applying a hard mask to those tokens in the attention map.
Finally, the restored attention output is added to the residual connection. This step is critical because the residual connection preserves the information of the unselected tokens from the previous layer.

As a result, Token Sparse Attention achieves dynamic token-level sparsification without sacrificing the expressive flexibility of dense attention. The model can selectively ignore irrelevant context to reduce computation, yet continuously re-evaluate token importance across layers and heads.
In the next subsection, we describe how token importance is estimated and how the per-head token subsets $S_H$ are dynamically determined under a sparsity budget.

\subsection{Dynamic Token Coverage}
\label{sec:dynamic_token_coverage}

\begin{algorithm}[t]
\caption{Dynamic Token Coverage and Index Search}
\label{alg:sparse_coverage}
\begin{algorithmic}[1]
\REQUIRE $Q$, $K$, Token Coverage $\tau$, Sequence Length $L$
\ENSURE Head-wise selected token indices $\{S_h\}_{h=1}^H$

\STATE \textcolor{gray}{\textit{// Compute attention scores using recent queries}}
\STATE $\hat{A} \leftarrow \mathrm{softmax}\left(Q_{[-last\_q:]}K^\top / \sqrt{d}\right)$

\STATE \textcolor{gray}{\textit{// Compute token-level scores per head}}
\STATE $s_h \leftarrow \mathrm{pooling}(\mathrm{sum}_{\text{vertical}}(\hat{A}))$

\STATE \textcolor{gray}{\textit{// Aggregate and normalize token scores across heads}}
\STATE $s_l \leftarrow (\sum_{h=1}^H s_{h})/ (\sum_{t=1}^L\sum_{h=1}^H s_{h}[t])$

\STATE \textcolor{gray}{\textit{// Sort tokens by ascending importance}}
\STATE $I \leftarrow \mathrm{argsort}(s_l)$

\STATE \textcolor{gray}{\textit{// Determine the number of sparse token}}
\STATE $k_{\text{sparse}} \leftarrow \arg\min_{k \in \{0,\dots,L\}}
\left\{\sum_{j=1}^{k} s_l[I[j]] \ge \tau \right\}$
\STATE $k_{\text{keep}} \leftarrow L - k_{\text{sparse}}$

\STATE \textcolor{gray}{\textit{// Head-wise top-$k_{\text{keep}}$ token selection}}
\STATE $\{S_h\}_{h=1}^H \leftarrow \{\mathrm{TopK}(s_h, k_{\text{keep}})\}_{h=1}^H$

\STATE \textbf{return} $\{S_h\}_{h=1}^H$
\end{algorithmic}
\end{algorithm}

Token Sparse Attention adaptively selects the sparsity budget at inference time, which involves two key decisions: (1) determining how many tokens to retain and (2) identifying which tokens to keep for each attention head. To address this, we introduce \textbf{Dynamic Token Coverage}, a token-selection policy applied before the compression stage of Token Sparse Attention. The full procedure is summarized in Algorithm~\ref{alg:sparse_coverage}.

We begin by estimating token importance independently for each attention head. For a given head $h$, we compute a lightweight proxy of the attention map $\hat{A}$ by attending a small set of recent queries to all keys. The importance score $s_h[t]$ for the $t$-th token is derived by summing the attention weights along the vertical axis (\textit{i.e.}, the sequence length dimension of queries). We implement this scoring step using a custom hardware-efficient kernel developed in Triton~\cite{triton} that minimizes memory I/O overhead by fusing the score calculation. 

\textbf{How Many Tokens to Retain.} Given the head-wise token scores, we aggregate them into a layer-level importance distribution by summing scores across heads and normalizing over the sequence dimension. This aggregated score captures the overall contribution of each token at the current layer and serves as the basis for determining the sparsity budget.

Instead of selecting tokens in descending order of importance, we sort tokens by ascending estimated importance and identify the minimal set of least important tokens whose cumulative mass exceeds a predefined coverage threshold $\tau$.
A key premise of our approach is that long-context attention accumulates attention noise, manifested as a long tail of tokens with negligible cumulative contribution. We interpret pruning this tail as a form of structural regularization that reduces distraction from irrelevant context.

\textbf{Which Tokens to Keep.} Once the layer-wise budget is determined, we perform the final token selection independently for each attention head by selecting the top-$k_{\text{keep}}$ tokens according to $s_h$. This design satisfies the heterogeneity requirement discussed in Section~\ref{sec:motivation}, allowing each head to attend to distinct semantic features while utilizing the allocated compute budget optimally.

\subsection{Sparse Layer Selection}
\label{sec:sparse_layer_selection}
Token Sparse Attention and Dynamic Token Coverage determine how tokens are selected within each layer. However, we observe that applying Token Sparse Attention across all layers leads to substantial performance degradation. This raises an orthogonal question: which layers can accommodate token-level sparsification with minimal impact on model behavior? 

\begin{figure}[t]
    \centering
    \includegraphics[width=1.0\linewidth]{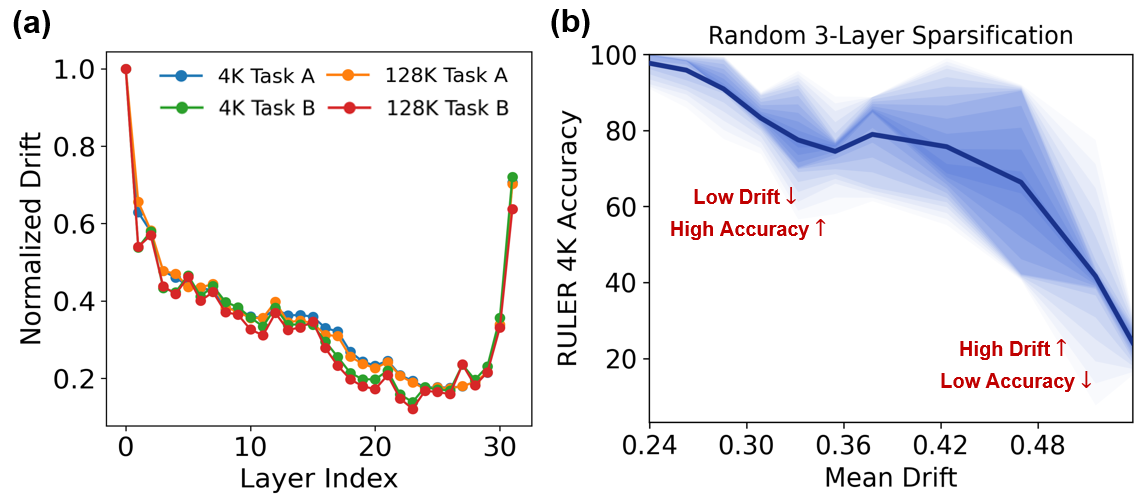}
     \vspace{-12pt}
    \caption{\textbf{Sparse Layer Selection.}
    \textbf{(a)} Layer-wise normalized drift measured across different tasks and context lengths. Task A and B correspond to a retrieval task and a summarization task, respectively. \textbf{(b)} Relationship between accuracy and drift under random 3-layer sparsification over 200 runs.
    }
    \vspace{-6pt}
    \label{fig:drift}
\end{figure}

To identify layers where token representations are sufficiently stable for sparsification, we introduce a metric called \textbf{Inter-Layer Representation Drift}, which measures the relative change in a token’s representation by comparing the L2 norms of its input and output hidden states. For a given layer $\ell$, the drift $R_\ell$ is defined as:
% \vspace{-6pt}
\begin{equation}
R_\ell = \mathbb{E}_t \left[ \frac{\lVert h_{\ell+1,t} - h_{\ell,t} \rVert_2}{\lVert h_{\ell,t} \rVert_2 + \epsilon} \right],
\end{equation}
where $h_{\ell,t}$ denotes the hidden state (\textit{i.e.}, layer input) of token $t$ at layer $\ell$. A lower drift value indicates smaller representational changes across layers, suggesting larger stability in token representations. As illustrated in Figure~\ref{fig:drift}\textbf{(a)}, the drift exhibits a consistent pattern across varying tasks and context lengths on LLaMA-3.1-8B-Instruct. 

\begin{table*}[t]
\centering
\caption{\textbf{Complementary Gains from Token Sparse Attention on RULER.} Adding Token Sparse Attention on top of FlashAttention and existing sparse attention methods improves speed with marginal accuracy variation across all context lengths and models. Speedup denotes the 128K attention speedup relative to the FlashAttention baseline.}
\label{tab:ruler}
\renewcommand{\arraystretch}{1.0}
\setlength{\tabcolsep}{7pt}
\scalebox{0.95}{
\begin{tabular}{lcccccccc}
\toprule
Method & 4K & 8K & 16K & 32K & 64K & 128K & Avg. & Speedup \\

\midrule
\multicolumn{9}{c}{\footnotesize LLaMA-3.1-8B-Instruct} \\
\midrule
Flash-Attention   & 95.82 & 92.77 & 91.02 & 84.87 & 83.43 & 74.15 & 87.01 & $\times$1.00 \\
\rowcolor{blue!5}
\textit{\quad w/ Token Sparse}  & 96.06 & 92.90 & 91.82 & 84.81 & 82.83 & 73.68 & 87.02 & $\times$1.36 \\
\midrule
Minference   & 93.46 & 92.29 & 91.01 & 85.34 & 83.19 & 73.63 & 86.49 & $\times$1.12 \\
\rowcolor{blue!5}
\textit{\quad w/ Token Sparse}  & 93.05 & 92.00 & 91.03 & 85.10 & 82.92 & 72.18 & 86.05 & $\times$1.38 \\
\midrule
FlexPrefill    & 95.48 & 92.71 & 91.40 & 87.20 & 83.05 & 73.75 & 87.27 & $\times$2.44 \\
\rowcolor{blue!5}
\textit{\quad w/ Token Sparse}  & 95.33 & 92.47 & 91.49 & 87.68 & 83.07 & 73.58 & 87.27 & $\times$2.76 \\

\midrule
\multicolumn{9}{c}{\footnotesize Mistral-Nemo-12B-Instruct} \\
\midrule
Flash-Attention   & 95.19 & 92.29 & 85.60 & 64.86 & 47.98 & 19.68 & 67.60 & $\times$1.00 \\
\rowcolor{blue!5}
\textit{\quad w/ Token Sparse}  & 95.07 & 92.10 & 85.97 & 63.67 & 47.97 & 19.44 & 67.37 & $\times$1.22 \\
\midrule
Minference   & 92.52 & 91.02 & 84.29 & 65.18 & 46.52 & 19.00 & 66.42 & $\times$1.13 \\
\rowcolor{blue!5}
\textit{\quad w/ Token Sparse}  & 93.01 & 91.36 & 84.31 & 64.70 & 46.67 & 18.68 & 66.46 & $\times$1.28 \\
\midrule
FlexPrefill   & 94.79 & 93.13 & 86.62 & 64.58 & 49.30 & 20.54 & 68.16 & $\times$1.22 \\
\rowcolor{blue!5}
\textit{\quad w/ Token Sparse}  & 94.84 & 93.31 & 86.14 & 64.89 & 48.70 & 19.58 & 67.91 & $\times$1.33 \\
\bottomrule
\end{tabular}
}
\vspace{-3pt}
\end{table*}

To investigate whether the drift is predictive of sparsification robustness, we conduct a controlled experiment shown in Figure~\ref{fig:drift}\textbf{(b)}. Specifically, we randomly sample three layers from the model and apply Token Sparse Attention only to these layers with token coverage of $0.99$, while keeping other layers dense. For each sampled triplet, we compute the mean normalized drift of the selected layers and evaluate the resulting accuracy on the RULER benchmark (4K-length). This process is repeated over 200 runs. The results reveal a clear correlation: layer subsets with lower average drift tend to exhibit higher accuracy, validating $R_\ell$ as a reliable indicator for sparsity robustness. Based on this observation, we use representation drift as a criterion to select layers for token-level sparsification.

We define a normalized drift rank for each layer and select the subset of layers eligible for Token Sparse Attention as:
% \vspace{-12pt}
\begin{equation}
\hat{R}_\ell = \frac{1}{L} \sum_{k=1}^{L} \mathbf{1}[R_k \leq R_\ell], \quad \mathcal{L}_{\text{sparse}}=\{\ell \mid \hat{R}_\ell \leq \delta\}
\end{equation}
In all our experiments, we set $\delta = 0.5$, applying our method strictly to the layers exhibiting the most stable representations. This layer selection is performed once as a preprocessing step for each model.

\begin{table*}[ht]
\centering
\caption{\textbf{Complementary Results from Token Sparse Attention on InfiniteBench.} }
%InfiniteBench results on LLaMA-3.1-8B-Instruct and Mistral-Nemo-12B-Instruct.
\vspace{-4pt}
\label{tab:infinitebench}

\renewcommand{\arraystretch}{1.05}
\setlength{\tabcolsep}{6pt}
\scalebox{0.90}{
\begin{tabular}{lccccccccc}
\toprule
Method & En.MC & En.QA & En.Dia & Retr.KV & Retr.N & Retr.P & Math.F & Code.D & Avg. \\
\midrule
\multicolumn{10}{c}{\footnotesize LLaMA-3.1-8B-Instruct} \\
\midrule
Flash-Attention & 64.19 & 27.64 & 19.00 & 55.20 & 98.47 & 97.80 & 24.00 & 20.56 & 50.86 \\
\rowcolor{blue!5}
\textit{\quad w/ Token Sparse} & 65.07 & 28.01 & 17.00 & 55.00 & 98.47 & 97.46 & 25.71 & 20.30 & 50.88 \\
\midrule
Minference & 67.25 & 26.80 & 18.50 & 51.60 & 97.46 & 97.80 & 22.29 & 19.54 & 50.16 \\
\rowcolor{blue!5}
\textit{\quad w/ Token Sparse} & 65.94 & 27.61 & 18.50 & 49.60 & 97.46 & 97.80 & 21.43 & 19.29 & 49.70 \\
\midrule
FlexPrefill & 68.12 & 27.37 & 11.00 & 50.80 & 99.15 & 96.95 & 22.57 & 20.30 & 49.53 \\
\rowcolor{blue!5}
\textit{\quad w/ Token Sparse} & 67.25 & 27.60 & 14.00 & 46.80 & 98.98 & 96.27 & 22.86 & 20.05 & 49.23 \\
\midrule
\multicolumn{10}{c}{\footnotesize Mistral-Nemo-12B-Instruct} \\
\midrule
Flash-Attention & 33.19 & 16.87 & 5.50 & 0.00 & 36.61 & 62.71 & 1.43 & 27.16 & 22.93 \\
\rowcolor{blue!5}
\textit{\quad w/ Token Sparse} & 33.19 & 16.81 & 5.50 & 0.00 & 35.76 & 62.88 & 0.86 & 27.41 & 22.80 \\
\midrule
Minference & 37.99 & 16.76 & 5.50 & 0.00 & 34.58 & 31.19 & 6.00 & 26.14 & 19.77 \\
\rowcolor{blue!5}
\textit{\quad w/ Token Sparse} & 38.43 & 16.30 & 5.50 & 0.00 & 33.73 & 30.00 & 5.14 & 26.14 & 19.41 \\
\midrule
FlexPrefill & 35.81 & 17.45 & 7.00 & 0.00 & 41.19 & 65.76 & 4.57 & 26.65 & 24.80 \\
\rowcolor{blue!5}
\textit{\quad w/ Token Sparse} & 35.37 & 16.71 & 6.50 & 0.00 & 39.32 & 65.93 & 3.14 & 25.63 & 24.08 \\
\bottomrule
\end{tabular}
}
\vspace{-4pt}
\end{table*}

\section{Experiments}

Our experiments first demonstrate the complementary gains achieved when Token Sparse Attention is combined with other attention-acceleration techniques in Section~\ref{sec:comparison_non_token}. We further compare it against token-eviction methods in Section~\ref{sec:comparison_token_eviction}. A detailed analysis of the accuracy-efficiency trade-off is provided in Section~\ref{sec:efficiency_results}.

\begin{figure*}[t]
    \centering
    \includegraphics[width=1.0\linewidth]{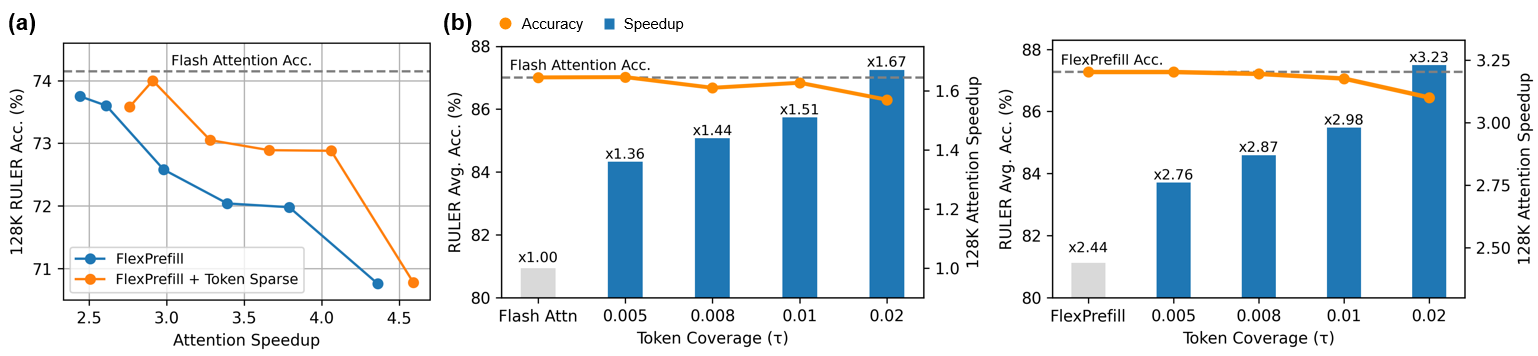}
     \vspace{-18pt}
    \caption{\textbf{Accuracy-Speedup Trade-offs.} \textbf{(a)} Accuracy-speedup Pareto frontier obtained by sweeping FlexPrefill hyperparameters, comparing with the Token Sparse Attention at token coverage of $\tau$=0.005. \textbf{(b)} Accuracy-speedup trade-off achieved by applying Token Sparse Attention with varying token coverage to FlashAttention (left) and FlexPrefill (right).
    }
    \vspace{-6pt}
    \label{fig:attention_acceleration}
\end{figure*}

\subsection{Experimental Setup}
\textbf{Models and Datasets.} We conduct experiments on LLaMA-3.1-8B-Instruct~\cite{llama3} and Mistral-Nemo-12B-Instruct~\cite{mistral}. For evaluation, we primarily use RULER~\cite{ruler} and InfiniteBench~\cite{infinitebench}, two benchmarks designed to assess long-context understanding and retrieval performance. Evaluation on LongBench~\cite{longbench} and Needle-in-a-Haystack~\cite{needle} is available in the Appendix~\ref{sec:appendix_experiments1}.

\textbf{Implementation Details.} Token Sparse Attention is integrated into FlashAttention without modifying the underlying kernel, and the token-scoring step in Section~\ref{sec:dynamic_token_coverage} is implemented using a Triton kernel~\cite{triton}. To balance efficiency and accuracy, we use a token-coverage parameter of $\tau{=}0.005$ for LLaMA and $\tau{=}0.008$ for Mistral across all experiments. All experiments are conducted on a single NVIDIA A100 80GB GPU.

\textbf{Baselines.} To validate our method's compatibility, several attention acceleration baselines are considered. FlashAttention~\cite{flashattention2} serves as the dense attention baseline, providing highly optimized exact attention with improved memory efficiency. Minference~\cite{minference} is a structured sparse attention method that applies predefined sparsity patterns to the attention map. We follow its official configuration, in which a Vertical-Slash sparsity pattern is applied uniformly across all attention heads. FlexPrefill~\cite{flexprefill} is a context-aware block-sparse attention method that dynamically prunes attention computation during prefill. Following the original paper, we set its hyperparameters to $\gamma{=}0.95$. To compare our method against token-eviction methods, PyramidInfer~\cite{pyramidinfer}, FastKV~\cite{fastkv} and GemFilter~\cite{gemfilter} are primarily considered. For fair comparison, all methods are applied only during prefill, while decoding uses standard dense attention. Additional experiments are provided in Appendix~\ref{sec:appendix_experiments2}.

\subsection{Accuracy Results}
\label{sec:comparison_non_token}
\textbf{RULER.} Table~\ref{tab:ruler} reports the complementary performance of several attention-acceleration methods with and without Token Sparse Attention. Across all context lengths, Token Sparse Attention largely preserves the accuracy of the underlying attention kernels across context lengths while improving the attention efficiency. For example, when applied to FlexPrefill on LLaMA-3.1-8B-Instruct, the average accuracy remains unchanged at 87.27\%, matching the vanilla FlexPrefill result.  A similar trend is observed when composing with FlashAttention and Minference. For Mistral-Nemo-12B-Instruct, the deviation introduced by Token Sparse Attention is consistently small, staying within 0.5\% of the baseline. In terms of efficiency, Token Sparse Attention reliably increases the attention speedup at 128K across all methods, \textit{e.g.}, for LLaMA-3.1-8B-Instruct, Minference improves from $\times1.12$ to $\times1.38$, and FlexPrefill from $\times2.44$ to $\times2.76$. Overall, Token Sparse Attention follows the baseline behavior closely, providing complementary efficiency gains with negligible impact on model accuracy.

\begin{figure*}[h]
    \centering
    \includegraphics[width=1.0\linewidth]{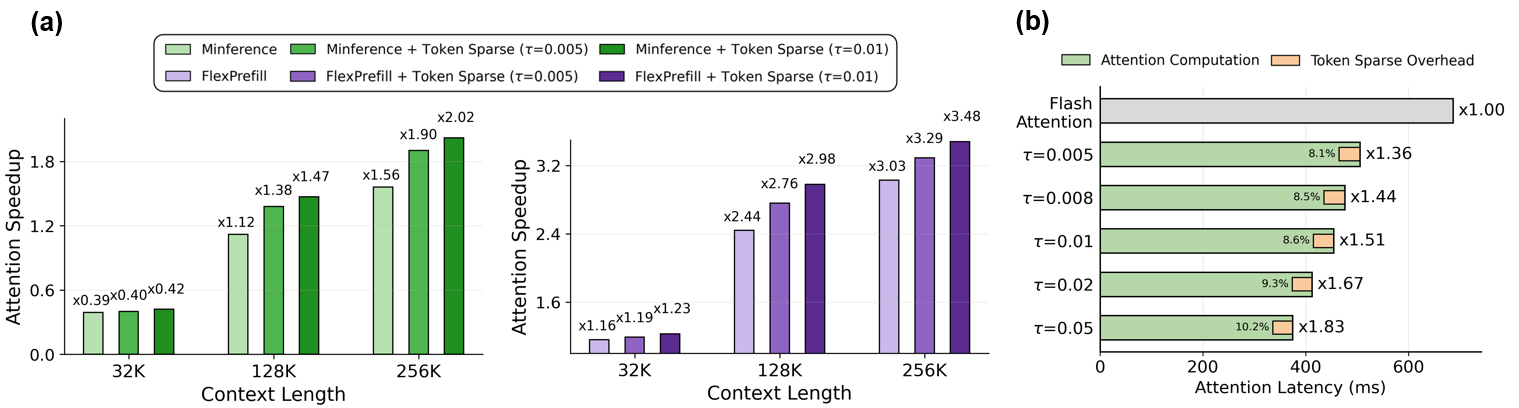}
     \vspace{-2pt}
    \caption{\textbf{Attention Speedup Details.} \textbf{(a)} Attention speedup across different context lengths. \textbf{(b)} Attention latency breakdown with Token Sparse Attention at 128K context, where the overhead includes token scoring and indexing, as well as QKV compression and attention output decompression.
    }
    \vspace{-8pt}
    \label{fig:seqlen_latency}
\end{figure*}

\textbf{InfiniteBench.} Table~\ref{tab:infinitebench} shows that Token Sparse Attention also preserves performance on InfiniteBench while remaining fully compatible with all baseline methods. For both LLaMA-3.1-8B-Instruct and Mistral-Nemo-12B-Instruct, augmenting Flash Attention, Minference, and FlexPrefill with Token Sparse Attention leads to only marginal accuracy differences, with overall results closely matching those of the corresponding baselines. The efficiency gains follow the same trend observed in Table~\ref{tab:ruler}. These results highlight that Token Sparse Attention acts as a general acceleration mechanism, providing consistent inference efficiency improvements while maintaining baseline accuracy.

\subsection{Efficiency Results}
\label{sec:efficiency_results}
\textbf{Accuracy-Speedup Trade-offs.}
We further analyze the computational benefits of Token Sparse Attention with respect to the accuracy–efficiency trade-offs. The attention speedup is estimated by the average attention latency measured across all layers. Figure~\ref{fig:attention_acceleration}\textbf{(a)} illustrates the Pareto frontier obtained by varying the hyperparameter $\gamma$, which represents the sparsity parameter, of FlexPrefill, a SOTA sparse attention (block-level) baseline. While aggressively tuning $\gamma$ yields higher speedups, it often comes at the cost of rapid accuracy degradation. In contrast, applying Token Sparse Attention (with $\tau{=}0.005$) on top of FlexPrefill consistently pushes the Pareto frontier outward, achieving superior speedups at comparable accuracy levels. This verifies that our method provides a complementary efficiency gain that cannot be achieved by simply adjusting hyperparameter.

Furthermore, we examine the effect of the token coverage parameter $\tau$ on performance. As shown in Figure~\ref{fig:attention_acceleration}\textbf{(b)}, increasing $\tau$ enables more aggressive token sparsification, yielding higher attention speedups for both FlashAttention and FlexPrefill. Remarkably, this aggressive reduction does not compromise model performance. The accuracy degradation remains within 1$\%$ even at higher sparsity levels. This robustness suggests that Token Sparse Attention effectively targets and removes only the irrelevant tokens in the context, allowing users to flexibly trade off a small margin of accuracy for significant gains in inference speed.

\begin{table}[t]
    \centering
    \caption{\textbf{Attention Map Sparsity.} Average attention map sparsity across different context lengths under different token coverage, $\tau$.}
    \label{tab:seqlen_sparsity}
    % \small
    \fontsize{8pt}{8pt}\selectfont
    \setlength{\tabcolsep}{3.5pt}
    \renewcommand{\arraystretch}{1.2}
    \begin{tabular}{lcccccc}
        \toprule
        \textbf{Sparsity} & \textbf{4K} & \textbf{8K} & \textbf{16K} & \textbf{32K} & \textbf{64K} & \textbf{128K} \\
        \midrule
        $\tau{=}0.005$ & 17.00\% & 21.11\% & 26.61\% & 28.44\% & 34.07\% & 54.44\% \\
        $\tau{=}0.010$ & 28.02\% & 32.98\% & 39.55\% & 41.25\% & 47.32\% & 67.36\% \\
        \bottomrule
    \end{tabular}
\end{table}

\textbf{Sparsity and Speedup across Sequence Lengths.}
Figure~\ref{fig:seqlen_latency}\textbf{(a)} shows that the attention speedup achieved by Token Sparse Attention increases consistently as the context length increases. While the gains are modest at shorter sequence lengths, substantially larger speedups are observed at 128K and 256K where attention computation dominates the overall latency. This behavior is explained by the increase in attention sparsity at longer contexts. As detailed in Table~\ref{tab:seqlen_sparsity}, the average attention sparsity within the selected layers increases steadily with sequence length for token coverage $\tau$. Together, these results demonstrate that Token Sparse Attention becomes increasingly effective in longer-context regimes, yielding larger efficiency gains when composed with existing attention baselines.

\textbf{Latency and Overhead Breakdown.}
Figure~\ref{fig:seqlen_latency}\textbf{(b)} presents a latency breakdown of Token Sparse Attention at 128K context length. Across all token coverage settings, the additional overhead introduced by Token Sparse Attention remains minimal, accounting for less than 11$\%$ of the total attention latency across all layers, even at the highest sparsity level. This overhead includes token scoring and indexing, as well as QKV compression and attention output decompression. These results confirm that Token Sparse Attention achieves significant acceleration while incurring only a small and well bounded overhead, validating its practicality for long context inference.

% \subsection{Analysis and Ablation Study}
\textbf{Dynamic Sparsity vs. Fixed Sparsity.} We compare our Dynamic Token Coverage against a fixed token sparsity baseline that keeps a constant fraction of tokens per layer. To ensure a fair comparison, we match the overall attention speedup at 128K by selecting fixed sparsity ratios $s$ that yield similar acceleration to dynamic sparsity. As shown in Table~\ref{tab:dynamic_fixed}, dynamic sparsity consistently achieves higher RULER average accuracy than fixed sparsity under comparable speedups. The gap becomes more pronounced at higher sparsity, where dynamic sparsity preserves accuracy substantially better than fixed sparsity while maintaining comparable acceleration. These results indicate that allocating the sparse budget based on the attention score distribution is more effective than enforcing a rigid token retention ratio, especially in long-context settings where token relevance varies significantly across inputs and layers.

\begin{table}[t]
\centering
% \small
\fontsize{8pt}{8pt}\selectfont
\caption{\textbf{Comparison between Dynamic and Fixed Sparsity.} $\tau$ denotes the token coverage used in dynamic sparsity, while $s$ represents the fixed token sparsity ratio. Sparsity is measured as the average attention map sparsity across the selected sparse layers. Experiments are based on RULER.}
\label{tab:dynamic_fixed}
\renewcommand{\arraystretch}{1.2}
\setlength{\tabcolsep}{8pt}
\begin{tabular}{lcccc}
\toprule
 & \multicolumn{2}{c}{\textbf{Dynamic Sparsity}} & \multicolumn{2}{c}{\textbf{Fixed Sparsity}} \\
 \cmidrule(lr){2-3} \cmidrule(lr){4-5}

\textbf{Metric} & $\tau{=}0.005$  & $\tau{=}0.010$  & $s{=}0.3$  & $s{=}0.5$ \\
\midrule
Sparsity & 54.44\% & 67.36\% & 50.96\% & 74.95\% \\
\midrule
Accuracy & 87.02\% & 86.84\% & 86.91\% & 85.43\% \\
Speedup & $\times$1.36 & $\times$1.51 & $\times$1.32 & $\times$1.57 \\
\bottomrule
\vspace{-8pt}
\end{tabular}
\end{table}

\subsection{Comparison with Token Eviction}
\label{sec:comparison_token_eviction}
We compare Token Sparse Attention with representative token eviction methods under a matched efficiency budget. Specifically, we set token coverage to $\tau=0.01$ and select the hyperparameter settings of PyramidInfer, FastKV and GemFilter that yield similar average attention speedup at 128K. As shown in Table~\ref{tab:token_eviction}, Token Sparse Attention attains the highest average RULER accuracy under comparable speedups. We attribute this advantage to our layer-wise dynamic budget allocation and reversible interleaving, where tokens skipped in one attention operation remain available via the residual path. In addition, head-wise selection enables fine-grained, head-specific token sets, avoiding the rigid unified token constraint of eviction-based methods.

\section{Related Works}
\textbf{Long-Context Inference for LLMs.} Handling long contexts in large language models introduces significant computational and memory challenges across different inference stages. During the prefill phase, the quadratic complexity of the attention operation leads to substantial computational overhead as the context length increases. To address these issues, system-level approaches such as FlashAttention~\cite{flashattention1,flashattention2,flashattention3} and FlashInfer~\cite{flashinfer} have been proposed to provide efficient attention frameworks. At the same time, algorithmic optimizations for long-context inference are actively being explored.

\textbf{Prefill Acceleration.} One major direction for prefill acceleration is token eviction and prompt compression. Methods such as FastKV~\cite{fastkv}, GemFilter~\cite{gemfilter}, and PyramidInfer~\cite{pyramidinfer} select important tokens at specific layers during the prefill phase and compress the corresponding hidden states to reduce computation. In contrast, prompt compression methods~\cite{prompt_comp1,prompt_comp2,prompt_comp3} operate at the prompt level rather than inside the model, reducing computation by shortening or transforming the input context itself.
Other works focus on sparse attention, which skips attention computation in certain regimes based on attention patterns. Minference~\cite{minference} applies sparse attention using a vertical-slash pattern, while FlexPrefill~\cite{flexprefill} introduces query-aware dynamic block-sparse attention. X-Attention~\cite{xattention} adopts an antidiagonal scoring strategy to reduce the search cost for identifying sparse patterns, whereas SeerAttention~\cite{seerattention} trains additional linear layers to search sparse attention patterns.
Model compression techniques such as quantization are also widely used. These methods reduce memory load time by compressing model weights or further compress activations to enable low-precision computation. Representative approaches include GPTQ~\cite{gptq}, SmoothQuant~\cite{smoothquant}, and AWQ~\cite{awq}. 

\textbf{Decoding Acceleration.} During the decoding phase, the computational bottleneck shifts from quadratic attention computation to the memory and bandwidth cost by KV cache. To address these challenges, methods such as H2O~\cite{h2o}, SnapKV~\cite{snapkv}, AdaKV~\cite{adakv}, and HeadKV~\cite{headkv} adopt KV cache eviction strategies that retain only important tokens in the KV cache during decoding. In contrast, approaches like ThinK~\cite{think} reduce decoding overhead by pruning channels of key representations rather than evicting tokens. In addition, applying quantization~\cite{kvquant,kivi,mikv} to keys and values to reduce KV cache loading time has also been actively explored. 

\begin{table}[t]
\centering
\caption{\textbf{Comparison against Token Eviction.} RULER accuracy comparison with token eviction methods on LLaMA-3.1-8B-Instruct. Speedups in the compared methods are set to a similar level, \textit{e.g.}, $\times$1.49 in PyramidInfer, $\times$1.50 in FastKV, $\times$1.53 in GemFilter, and $\times$1.51 in Ours, compared to FlashAttn.}
\label{tab:token_eviction}
% \scriptsize
\fontsize{8pt}{8pt}\selectfont
\setlength{\tabcolsep}{4.5pt}  
\renewcommand{\arraystretch}{1.2}
\begin{tabular}{lccccccc}
\toprule
\textbf{Method} & \textbf{4K} & \textbf{8K} & \textbf{16K} & \textbf{32K} & \textbf{64K} & \textbf{128K} & \textbf{Avg.} \\
\midrule
FlashAttn & 95.82 & 92.77 & 91.02 & 84.87 & 83.43 & 74.15 & 87.01 \\
\midrule
PyramidInfer    & 83.15 & 79.27 & 80.74 & 80.42 & 77.98 & 69.35 & 78.49 \\
GemFilter       & 92.50 & 90.87 & 88.78 & 85.01 & 80.42 & 73.15 & 85.12 \\
FastKV          & 94.25 & 91.30 & 89.78 & 84.57 & 81.54 & 72.39 & 85.64 \\

\rowcolor{blue!5}
Ours    & 95.82 & 92.71 & 91.46 & 84.81 & 83.00 & 73.25 & 86.84 \\
\bottomrule
\end{tabular}
\end{table}

\section{Conclusion}
In this paper, we introduced Token Sparse Attention, a dynamic and reversible token-level sparsification mechanism for efficient long-context inference. The key advantage of Token Sparse Attention lies in its ability to reduce attention computation without permanently removing tokens, allowing token relevance to be re-evaluated across layers and heads. Moreover, our design is fully compatible with existing dense and sparse attention kernels, enabling seamless composition with prior acceleration methods.
Experimental results on long-context benchmarks demonstrate that Token Sparse Attention consistently improves the accuracy--latency trade-off across models and tasks. By selectively filtering irrelevant tokens before attention computation, our method achieves substantial attention speedups with minimal accuracy degradation.
Ultimately, Token Sparse Attention provides a complementary and practical solution for scalable long-context inference, offering an effective way to improve efficiency while preserving model behavior.

\section*{Impact Statement}
This paper presents work whose goal is to advance the field of Machine
Learning. There are many potential societal consequences of our work, none
which we feel must be specifically highlighted here.

\section*{Acknowledgements}
This work was supported in part by Institute of Information $\&$ communications Technology Planning $\&$ Evaluation (IITP) grant funded by the Korea government (MSIT) (No.RS-2025-02273157: Development of Low Power Training/Inference Technologies based on AI Semiconductors, No.RS-2026-25507427: Development of Efficient Architectures and Training Techniques for High-Performance Lightweight AI Models, No. 2021-0-01343: Artificial Intelligence Graduate School Program (Seoul National University)), Samsung Research Funding Center under Project SRFC-TC1603-53, BK21 FOUR program, and the Sejong Fellowship of National Research Foundation of Korea (NRF) funded by the Korea government (MSIT) (No. RS-2025-00561953: Efficient Deep Learning and Inference Algorithms for On-Device AI). (Corresponding Authors: Beomseok Kang and Jae-Joon Kim).

\nocite{*}

\bibliography{reference}
\bibliographystyle{icml2026}

%%%%%%%%%%%%%%%%%%%%%%%%%%%%%%%%%%%%%%%%%%%%%%%%%%%%%%%%%%%%%%%%%%%%%%%%%%%%%%%
%%%%%%%%%%%%%%%%%%%%%%%%%%%%%%%%%%%%%%%%%%%%%%%%%%%%%%%%%%%%%%%%%%%%%%%%%%%%%%%
% APPENDIX
%%%%%%%%%%%%%%%%%%%%%%%%%%%%%%%%%%%%%%%%%%%%%%%%%%%%%%%%%%%%%%%%%%%%%%%%%%%%%%%
%%%%%%%%%%%%%%%%%%%%%%%%%%%%%%%%%%%%%%%%%%%%%%%%%%%%%%%%%%%%%%%%%%%%%%%%%%%%%%%
\newpage
\appendix
\onecolumn
\section{Appendix}
\subsection{Additional Models and Datasets Details}
\textbf{Models.}
To evaluate our approach in the long-context regime, we conduct experiments on LLaMA-3.1-8B-Instruct~\cite{llama3} and Mistral-Nemo-12B-Instruct~\cite{mistral}, designed to operate with 128K context window.

\sloppy

The checkpoints for all models are publicly available at:

\noindent\textbf{LLaMA-3.1-8B-Instruct}: \\
\url{https://huggingface.co/meta-llama/Llama-3.1-8B-Instruct} \\
\textbf{Mistral-Nemo-12B-Instruct}:\\
\url{https://huggingface.co/mistralai/Mistral-Nemo-Instruct-2407} \\

\textbf{RULER.}
RULER is a configurable synthetic benchmark designed to assess the long-context capabilities of large language models under varying sequence lengths and task difficulties. Rather than focusing on a single retrieval scenario, it generalizes the Needle-in-a-Haystack setup into a comprehensive benchmark suite comprising 13 tasks grouped into four categories: retrieval-style tasks, aggregation tasks (e.g., CWE and FWE), multi-hop tracing tasks (VT), and question answering. This diverse task design allows for controlled and systematic evaluation of a model’s ability to retrieve, aggregate, and reason over long contexts beyond simple keyword matching.

\textbf{InfiniteBench}
InfiniteBench is a large-scale benchmark designed to evaluate the ability of language models to understand and reason over extremely long contexts that exceed 100K tokens. It includes a diverse collection of tasks spanning multiple domains, such as retrieval, reasoning, code understanding, mathematical computation, dialogue, and summarization, and covers both synthetic and realistic scenarios.

\subsection{Additional Benchmarks}
\label{sec:appendix_experiments1}
\begin{figure}[h]
    \centering
    \includegraphics[width=0.95\linewidth]{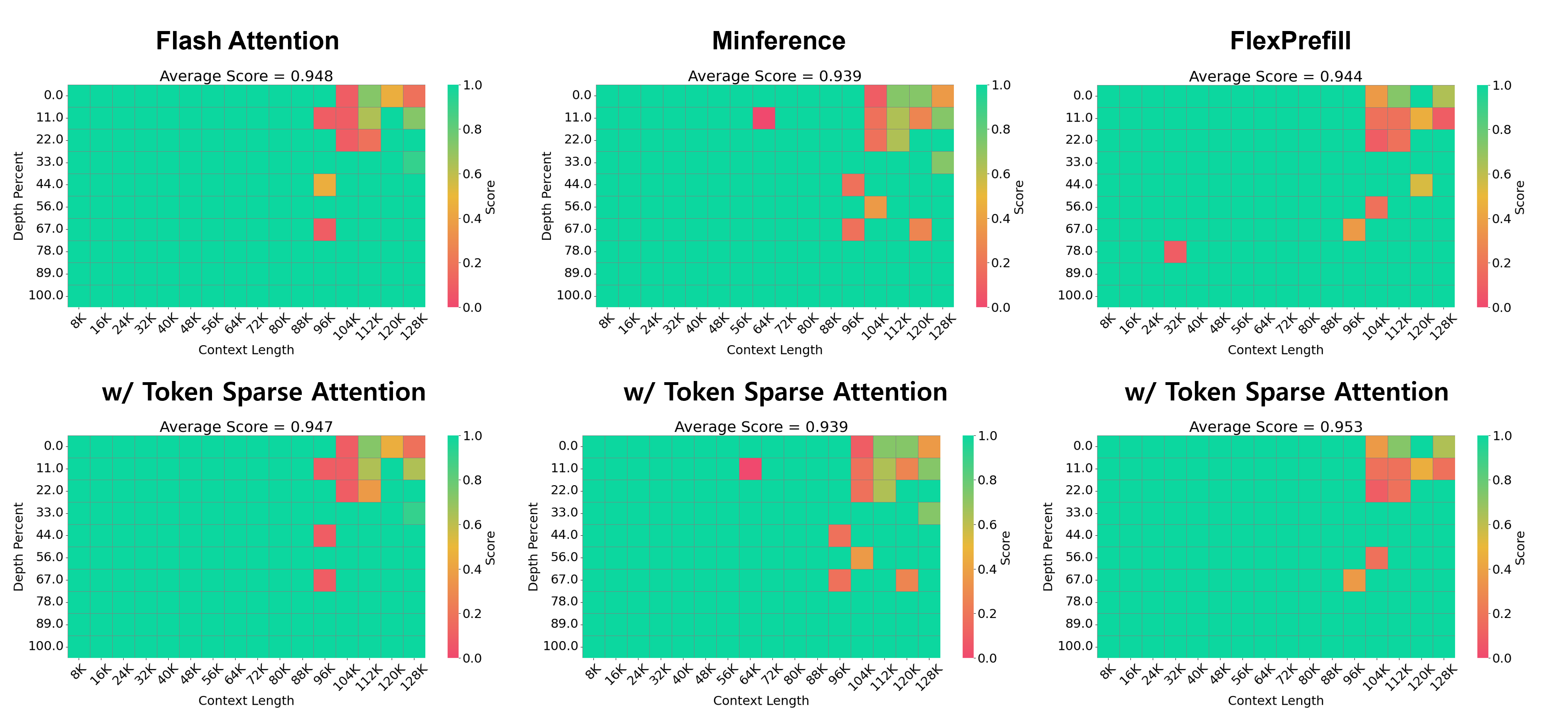}
     \vspace{-2pt}
    \caption{
    Needle-in-a-Haystack results with Token Sparse Attention on LLaMA-3.1-8B-Instruct.
    }
    \vspace{-8pt}
    \label{fig:niah}
\end{figure}

\textbf{Needle-in-a-Haystack} We provide Needle-in-a-Haystack results for LLaMA-3.1-8B-Instruct in Figure~\ref{fig:niah}.
Token Sparse Attention demonstrates strong accuracy preservation under long-context settings. When composed with FlexPrefill, Token Sparse Attention consistently achieves higher accuracy than FlexPrefill alone, These results suggest that reversible token-level sparsification effectively removes irrelevant context while preserving the critical signal required for precise retrieval in long-context scenarios.

\newpage

\textbf{LongBench.} Table~\ref{tab:longbench} reports the LongBench results on LLaMA-3.1-8B-Instruct. LongBench covers a broad spectrum of long-context tasks, including single-document and multi-document question answering, summarization, few-shot learning, synthetic reasoning, and code understanding, providing a comprehensive evaluation of model robustness under long-context settings. Across all baselines, Token Sparse Attention preserves performance consistently while introducing minimal accuracy changes.

\newcommand{\rothead}[1]{\rotatebox{60}{#1}}
\begin{table*}[h]

\centering
\setlength{\tabcolsep}{3pt}
\renewcommand{\arraystretch}{1.1}
\caption{LongBench results with Token Sparse Attention on LLaMA-3.1-8B-Instruct.}
\label{tab:longbench}
\scalebox{0.70}{
\begin{tabular}{l|ccc|ccc|ccc|ccc|cc|cc|c}
\toprule
 & \multicolumn{3}{c|}{Single-Document QA}
 & \multicolumn{3}{c|}{Multi-Document QA}
 & \multicolumn{3}{c|}{Summarization}
 & \multicolumn{3}{c|}{Few-shot Learning}
 & \multicolumn{2}{c|}{Synthetic}
 & \multicolumn{2}{c|}{Code}
 &  \\
Method
        & \rothead{NrtvQA}
        & \rothead{Qasper}
        & \rothead{MF-en}
        & \rothead{HotpotQA}
        & \rothead{2WikiMQA}
        & \rothead{MuSiQue}
        & \rothead{GovReport}
        & \rothead{QMSum}
        & \rothead{MultiNews}
        & \rothead{TREC}
        & \rothead{TriviaQA}
        & \rothead{SAMSum}
        & \rothead{LCC}
        & \rothead{RB-P}
        & \rothead{PCount}
        & \rothead{PRe}
        & Avg. \\
\midrule
Flash-Attention
& 30.22 & 45.37 & 55.80
& 55.97 & 45.00 & 31.26
& 35.12 & 25.38 & 27.20
& 72.50 & 91.65 & 44.32
& 9.41 & 99.50
& 62.90 & 56.80
& 49.28 \\
\rowcolor{blue!5}
\textit{\quad w/ Token Sparse}
& 30.43 & 45.17 & 55.34
& 55.00 & 45.56 & 31.28
& 34.97 & 25.43 & 27.09
& 72.50 & 91.64 & 44.07
& 7.47 & 98.00
& 63.06 & 56.28
& 48.96 \\
\midrule
Minference
& 29.48 & 45.64 & 52.59
& 55.05 & 42.93 & 29.24
& 34.88 & 24.94 & 26.53
& 71.00 & 91.89 & 44.15
& 6.38 & 98.50
& 62.38 & 50.72
& 47.89 \\
\rowcolor{blue!5}
\textit{\quad w/ Token Sparse}
& 29.28 & 46.19 & 52.52
& 54.62 & 42.18 & 29.89
& 34.79 & 25.05 & 26.48
& 71.00 & 91.89 & 44.07
& 5.34 & 94.50
& 62.79 & 51.12
& 47.61 \\
\midrule
FlexPrefill
& 27.99 & 44.63 & 54.94
& 57.31 & 41.82 & 31.97
& 34.50 & 24.92 & 27.06
& 70.50 & 90.74 & 43.81
& 4.09 & 82.00
& 63.08 & 60.36
& 47.48 \\
\rowcolor{blue!5}
\textit{\quad w/ Token Sparse}
& 28.07 & 44.84 & 54.74
& 56.63 & 40.24 & 31.09
& 34.79 & 25.47 & 27.12
& 69.00 & 90.89 & 43.72
& 3.09 & 85.50
& 62.78 & 60.24
& 47.39 \\
\bottomrule
\end{tabular}
}
\vspace{-2pt}
\end{table*}

\subsection{Additional Ablation Study}

\textbf{Token Scoring Methods.} To validate our scoring design, we conduct additional ablation experiments comparing three alternatives: random query selection, query-only pooling, and our recent-query approach. Random query selection computes importance scores using randomly sampled queries instead of recent queries. Query-only pooling compresses the query tensor by pooling along the sequence dimension and computes approximate attention scores against all keys using the resulting compressed representation.

As shown in Table~\ref{tab:scoring_abl}, The results confirm that random query selection leads to noticeable accuracy degradation, and query-only pooling partially recovers accuracy but still falls short of recent-query scoring. These results validate that recent-query scoring is an effective design choice for token-level importance estimation, as it leverages the most contextually relevant queries to identify important tokens without incurring significant computational overhead.

\begin{table*}[h]
\centering
\caption{Ablation study on scoring methods with Token Sparse Attention on RULER using LLaMA-3.1-8B-Instruct.}
\label{tab:scoring_abl}
\renewcommand{\arraystretch}{1.0}
\setlength{\tabcolsep}{7pt}
\scalebox{0.95}{
\begin{tabular}{llccccccc}
\toprule
Method & Scoring & 4K & 8K & 16K & 32K & 64K & 128K & Avg. \\
\midrule
Flash-Attention & -- & 95.82 & 92.77 & 91.02 & 84.87 & 83.43 & 74.15 & 87.01 \\
\textit{\quad w/ Token Sparse} & Random Query   & 94.28 & 92.49 & 90.05 & 82.38 & 80.83 & 69.66 & 84.95 \\
\textit{\quad w/ Token Sparse} & Only Q-Pooling & 95.02 & 92.55 & 91.35 & 84.68 & 83.04 & 71.92 & 86.43 \\
\textit{\quad w/ Token Sparse} & Recent-Query   & 96.06 & 92.90 & 91.82 & 84.81 & 82.83 & 73.68 & 87.02 \\
\bottomrule
\end{tabular}
}
\vspace{6pt}
\end{table*}

\textbf{Stability of Drift-Based Layer Selection.} To validate the effectiveness of the drift-based layer selection policy introduced in Section~\ref{sec:sparse_layer_selection}, we conduct an ablation study examining how the choice of sparse layers affects model accuracy. Specifically, we partition all layers into three equal groups based on their normalized drift magnitude (Low, Mid, and High, each comprising the bottom, middle, and top 33$\%$ of layers by drift value) and apply Token Sparse Attention exclusively to the layers within each group.

Table~\ref{tab:drift_abl} reports the mean drift and RULER accuracy for each group across both models. Applying Token Sparse Attention to Low Drift layers consistently yields the highest accuracy, while restricting sparsification to High Drift layers leads to substantial degradation. This trend holds across both LLaMA-3.1-8B-Instruct and Mistral-Nemo-12B-Instruct, demonstrating that drift is a reliable and model-agnostic indicator for identifying layers that can accommodate token-level sparsification with minimal impact on accuracy.

\newpage

\begin{table*}[h]
\centering
\caption{RULER Accuracy under drift-based layer grouping for Token Sparse Attention.}
\label{tab:drift_abl}
\renewcommand{\arraystretch}{1.0}
\setlength{\tabcolsep}{7pt}
\scalebox{0.95}{
\begin{tabular}{lcccccccc}
\toprule
Sparse Layer & Mean Drift & 4K & 8K & 16K & 32K & 64K & 128K & Avg. \\
\midrule
\multicolumn{9}{c}{\footnotesize LLaMA-3.1-8B-Instruct} \\
\midrule
Low Drift  & 0.204 & 95.82 & 92.52 & 90.95 & 84.87 & 83.19 & 73.99 & 86.89 \\
Mid Drift  & 0.348 & 95.38 & 92.28 & 91.40 & 83.32 & 81.21 & 69.48 & 85.51 \\
High Drift & 0.553 & 94.67 & 91.74 & 89.08 & 68.88 & 50.24 & 38.17 & 72.13 \\
\midrule
\multicolumn{9}{c}{\footnotesize Mistral-Nemo-12B-Instruct} \\
\midrule
Low Drift  & 0.088 & 95.13 & 92.22 & 85.30 & 64.26 & 46.71 & 19.30 & 67.15 \\
Mid Drift  & 0.146 & 95.17 & 91.74 & 81.09 & 66.73 & 42.45 & 18.64 & 65.97 \\
High Drift & 0.299 & 70.92 & 29.57 & 24.93 & 17.43 & 6.41  & 4.03  & 25.55 \\
\bottomrule
\end{tabular}
}
\vspace{2pt}
\end{table*}

\textbf{Delta Sweep for Sparse Layer Selection.} Table~\ref{tab:delta_abl} reports the accuracy-speedup trade-off across varying delta values on RULER for LLaMA-3.1-8B-Instruct. The layer selection threshold delta controls the proportion of layers eligible for token-level sparsification, where a larger delta includes more layers with higher representation drift. The results demonstrate that $\delta$=0.5 serves as a consistent sweet spot, balancing accuracy preservation and speedup. Values beyond 0.5 yield only marginal additional speedup while incurring notable accuracy degradation.

\begin{table*}[h]
\centering
\caption{Effect of delta on Token Sparse Attention on RULER using LLaMA-3.1-8B-Instruct.}
\label{tab:delta_abl}
\renewcommand{\arraystretch}{1.0}
\setlength{\tabcolsep}{7pt}
\scalebox{0.95}{
\begin{tabular}{lccccccccc}
\toprule
Method & $\delta$ & 4K & 8K & 16K & 32K & 64K & 128K & Avg. & Speedup \\
\midrule
Flash-Attention & -- & 95.82 & 92.77 & 91.02 & 84.87 & 83.43 & 74.15 & 87.01 & $\times$1.00 \\
\textit{\quad w/ Token Sparse} & 0.1 & 95.82 & 92.47 & 91.16 & 84.95 & 83.02 & 73.94 & 86.89 & $\times$1.12 \\
\textit{\quad w/ Token Sparse} & 0.3 & 95.82 & 92.71 & 91.00 & 84.87 & 83.19 & 73.91 & 86.92 & $\times$1.24 \\
\textit{\quad w/ Token Sparse} & 0.5 & 96.06 & 92.90 & 91.82 & 84.81 & 82.83 & 73.68 & 87.02 & $\times$1.36 \\
\textit{\quad w/ Token Sparse} & 0.7 & 95.30 & 92.62 & 90.58 & 83.44 & 81.53 & 68.63 & 85.35 & $\times$1.46 \\
\textit{\quad w/ Token Sparse} & 0.9 & 94.68 & 89.94 & 90.23 & 73.14 & 70.55 & 56.41 & 79.16 & $\times$1.53 \\
\bottomrule
\end{tabular}
}
\vspace{-3pt}
\end{table*}

\subsection{Prefill Latency}

Our method and baselines target prefill acceleration, and since all methods use dense attention with full KV cache during decoding, TPOT is identical across all methods (89ms at 128K with 128 generation tokens, measured on FlashAttention). End-to-end gains are therefore captured through prefill latency. To further support this, we additionally report full prefill latency (TTFT) on both A100 and A6000 GPUs.

As shown in Table~\ref{tab:prefill_latency_a100} and Table~\ref{tab:prefill_latency_a6000}, at 128K, prefill latency improvements closely follow attention-level speedups on both hardware platforms, confirming that attention dominates in long-context regimes. At 8K, the attention sparsity decreases as shown in Table~\ref{tab:seqlen_sparsity}, and the scoring overhead becomes relatively more visible. However, the overhead introduced by Token Sparse Attention remains small compared to other baselines, and our primary goal is to push the efficiency frontier in long-context inference, where consistent gains are observed across all methods and both hardware configurations.

\begin{table*}[h]
\centering
\caption{Prefill latency comparison across 8K and 128K sequence lengths on a single A100 GPU}
\label{tab:prefill_latency_a100}
\renewcommand{\arraystretch}{1.0}
\setlength{\tabcolsep}{10pt}
\scalebox{0.95}{
\begin{tabular}{lcccc}
\toprule
& \multicolumn{2}{c}{8K} & \multicolumn{2}{c}{128K} \\
\cmidrule(lr){2-3} \cmidrule(lr){4-5}
Method & Latency (sec) & Speedup & Latency (sec) & Speedup \\
\midrule
Flash-Attention & $0.68 \pm 0.0061$ & $\times$1.00 & $31.04 \pm 0.0569$ & $\times$1.00 \\
\rowcolor{blue!5}
\textit{\quad w/ Token Sparse} & $0.70 \pm 0.0064$ & $\times$0.97 & $24.35 \pm 0.0195$ & $\times$1.27 \\
\midrule
Minference & $1.52 \pm 0.0079$ & $\times$0.45 & $26.84 \pm 0.0492$ & $\times$1.16 \\
\rowcolor{blue!5}
\textit{\quad w/ Token Sparse} & $1.53 \pm 0.0103$ & $\times$0.44 & $23.32 \pm 0.0124$ & $\times$1.33 \\
\midrule
FlexPrefill & $0.90 \pm 0.0084$ & $\times$0.76 & $18.45 \pm 0.0439$ & $\times$1.68 \\
\rowcolor{blue!5}
\textit{\quad w/ Token Sparse} & $0.92 \pm 0.0098$ & $\times$0.74 & $16.81 \pm 0.0272$ & $\times$1.85 \\
\bottomrule
\end{tabular}
}
\vspace{-3pt}
\end{table*}

\begin{table*}[h]
\centering
\caption{Prefill latency comparison across 8K and 128K sequence lengths on a single A6000 GPU}
\label{tab:prefill_latency_a6000}
\renewcommand{\arraystretch}{1.0}
\setlength{\tabcolsep}{10pt}
\scalebox{0.95}{
\begin{tabular}{lcccc}
\toprule
& \multicolumn{2}{c}{8K} & \multicolumn{2}{c}{128K} \\
\cmidrule(lr){2-3} \cmidrule(lr){4-5}
Method & Latency (sec) & Speedup & Latency (sec) & Speedup \\
\midrule
Flash-Attention & $1.76 \pm 0.3624$ & $\times$1.00 & $67.34 \pm 0.4273$ & $\times$1.00 \\
\rowcolor{blue!5}
\textit{\quad w/ Token Sparse} & $2.04 \pm 0.2284$ & $\times$0.86 & $54.49 \pm 0.1430$ & $\times$1.24 \\
\midrule
Minference & $4.00 \pm 0.5044$ & $\times$0.44 & $53.54 \pm 0.6051$ & $\times$1.26 \\
\rowcolor{blue!5}
\textit{\quad w/ Token Sparse} & $3.71 \pm 0.4937$ & $\times$0.47 & $47.05 \pm 0.3972$ & $\times$1.43 \\
\midrule
FlexPrefill & $3.22 \pm 0.5334$ & $\times$0.55 & $47.16 \pm 0.3978$ & $\times$1.43 \\
\rowcolor{blue!5}
\textit{\quad w/ Token Sparse} & $2.68 \pm 0.3400$ & $\times$0.66 & $43.08 \pm 0.4519$ & $\times$1.56 \\
\bottomrule
\end{tabular}
}
\vspace{-3pt}
\end{table*}

\subsection{Comparison with Additional Baselines}
\label{sec:appendix_experiments2}
To further validate the complementary nature of Token Sparse Attention, we conduct additional experiments composing it with SeerAttention~\cite{seerattention} and X-Attention~\cite{xattention} on LLaMA-3.1-8B-Instruct. Both methods apply structured sparsity patterns at the attention map level, and Token Sparse Attention operates orthogonally at the Q, K, V tensor level, enabling seamless composition without modifying the underlying sparse kernels.

As shown in Table~\ref{tab:x-attn_seerattn}, composing Token Sparse Attention with SeerAttention improves the attention speedup from x2.19 to x2.47 while maintaining comparable accuracy. Similarly, when composed with X-Attention (S=16, $\tau$=0.9), the speedup increases from x2.72 to x3.49 with negligible accuracy change. These results are consistent with the complementary gains observed with MInference and FlexPrefill in the main experiments, and suggest that Token Sparse Attention provides a general acceleration benefit that extends across diverse sparse attention kernels regardless of their specific sparsity pattern.

\begin{table*}[h]
\centering
\caption{Complementary Gains of additional methods from Token Sparse Attention on RULER.}
\label{tab:x-attn_seerattn}
\renewcommand{\arraystretch}{1.0}
\setlength{\tabcolsep}{7pt}
\scalebox{0.95}{
\begin{tabular}{lcccccccc}
\toprule
Method & 4K & 8K & 16K & 32K & 64K & 128K & Avg. & Speedup \\

\midrule
\multicolumn{9}{c}{\footnotesize LLaMA-3.1-8B-Instruct} \\
\midrule
Flash-Attention   & 95.82 & 92.77 & 91.02 & 84.87 & 83.43 & 74.15 & 87.01 & $\times$1.00 \\
\rowcolor{blue!5}
\textit{\quad w/ Token Sparse}  & 96.06 & 92.90 & 91.82 & 84.81 & 82.83 & 73.68 & 87.02 & $\times$1.36 \\
\midrule
SeerAttention   & 95.82 & 92.75 & 90.97 & 84.81 & 83.71 & 73.85 & 86.99 & $\times$2.19 \\
\rowcolor{blue!5}
\textit{\quad w/ Token Sparse}  & 95.82 & 92.84 & 91.47 & 85.17 & 83.19 & 73.60 & 87.02 & $\times$2.47 \\
\midrule
X-Attention   & 96.14 & 92.46 & 89.99 & 84.05 & 76.05 & 61.65 & 83.39 & $\times$2.72 \\
\rowcolor{blue!5}
\textit{\quad w/ Token Sparse}  & 96.14 & 92.16 & 90.01 & 84.26 & 76.03 & 62.43 & 83.51 & $\times$3.49 \\
\bottomrule
\end{tabular}
}
\vspace{-3pt}
\end{table*}

\subsection{Comparison with Non-Eviction Method}

We compare Token Sparse Attention against OrthoRank~\cite{orthorank}, a non-eviction method that identifies important tokens based on their orthogonality to the sink token in the normalized hidden state space. Although OrthoRank avoids permanent token removal, we note two fundamental architectural differences that distinguish it from Token Sparse Attention.

\textbf{Compression Level and Scoring Mechanism.} OrthoRank operates at the layer input hidden state level, selecting tokens prior to the attention module based on a geometric metric derived from the normalized hidden state space. In contrast, Token Sparse Attention performs compression at the Q, K, V level within each attention head, using the attention map itself as the scoring signal. Since our goal is to identify which tokens are important specifically for the attention computation, using the attention map as a scoring proxy directly captures the degree to which each token contributes to the attention output at the current layer and head.

\textbf{Per-Head Independent Token Selection.} OrthoRank selects tokens based on the hidden state prior to the attention module, which means all attention heads within the same layer share an identical token set. This design is unable to account for the head-wise divergence in attention patterns. As shown in Figure~\ref{fig:motivation}(b), different attention heads within the same layer exhibit significantly different token importance rankings, and enforcing a single unified token set across all heads inevitably discards tokens that certain heads would find essential. 

To empirically validate these differences, we reproduced OrthoRank and extended it to the long-context evaluation setting. For a fair comparison, we use prefill speedup rather than attention speedup as the efficiency metric, since OrthoRank additionally skips FFN computation for unselected tokens, making attention speedup alone an insufficient basis for comparison.

As shown in Table~\ref{tab:orthorank}, Token Sparse Attention achieves substantially higher average RULER accuracy (87.02$\%$ and 86.84$\%$) compared to OrthoRank at both pruning ratios (79.36 $\%$ and 78.38$\%$) under comparable prefill speedup. The accuracy gap is particularly pronounced at longer context lengths, with a difference of approximately 17$\%$ at the 128K setting. We attribute this gap primarily to the two architectural differences described above.

\begin{table*}[h]
\centering
\caption{RULER accuracy comparison with non-eviction method on LLaMA-3.1-8B-Instruct. $p$ denotes the pruning ratio.}
\label{tab:orthorank}
\renewcommand{\arraystretch}{1.0}
\setlength{\tabcolsep}{7pt}
\scalebox{0.95}{
\begin{tabular}{lcccccccc}
\toprule
Method & 4K & 8K & 16K & 32K & 64K & 128K & Avg. & Prefill Speedup \\

\midrule
\multicolumn{9}{c}{\footnotesize LLaMA-3.1-8B-Instruct} \\
\midrule
Flash-Attention   & 95.82 & 92.77 & 91.02 & 84.87 & 83.43 & 74.15 & 87.01 & $\times$1.00 \\
\midrule
OrthoRank ($p$=0.30)              & 90.25 & 89.85 & 86.37 & 81.10 & 72.60 & 55.98 & 79.36 & $\times$1.27 \\
OrthoRank ($p$=0.35)              & 90.86 & 86.23 & 84.54 & 80.89 & 71.36 & 56.39 & 78.38 & $\times$1.31 \\
Token Sparse ($\tau$=0.005)       & 96.06 & 92.90 & 91.82 & 84.81 & 82.83 & 73.68 & 87.02 & $\times$1.27 \\
Token Sparse ($\tau$=0.010)       & 95.82 & 92.71 & 91.46 & 84.81 & 83.00 & 73.25 & 86.84 & $\times$1.32 \\
\bottomrule
\end{tabular}
}
\vspace{-3pt}
\end{table*}

\subsection{Limitation}
While Token Sparse Attention is primarily designed for long-context inference, its efficiency gains are naturally smaller at short context lengths where attention computation does not yet dominate end-to-end latency. This behavior is common to most coverage-based sparse attention methods and does not detract from the effectiveness of our approach in the long-context regimes that are the main focus of this work. Meanwhile, applying overly aggressive token sparsification can increase the risk of excluding semantically important tokens from attention, potentially leading to accuracy degradation. However, Token Sparse Attention is designed to remove irrelevant tokens that contribute primarily to attention noise rather than informative context. By selecting an appropriate token coverage, the method enables a controlled and practical accuracy–efficiency trade-off, which is the primary objective in long-context regimes. Finally, while this work focuses on prefill acceleration for text generation models, Token Sparse Attention can be extended in several promising directions. Future work includes adapting the proposed mechanism to the decoding phase, as well as exploring its applicability to multimodal settings such as vision–language models.

%%%%%%%%%%%%%%%%%%%%%%%%%%%%%%%%%%%%%%%%%%%%%%%%%%%%%%%%%%%%%%%%%%%%%%%%%%%%%%%
%%%%%%%%%%%%%%%%%%%%%%%%%%%%%%%%%%%%%%%%%%%%%%%%%%%%%%%%%%%%%%%%%%%%%%%%%%%%%%%

\end{document}